\documentclass{article}
\usepackage{graphicx} 
\usepackage{amsmath, amssymb, amsthm, mathtools, bm}
\usepackage{float}
\usepackage[colorlinks=true, allcolors=blue]{hyperref}
\usepackage{algorithm}
\usepackage{todonotes}
\usepackage{algpseudocode}
\usepackage{thmtools, thm-restate}
\usepackage{listings}

\declaretheorem[name=Theorem]{theorem}

\declaretheorem[name=Lemma]{lemma}

\title{Iterative Orthogonalization Scaling Laws}
\author{
Devan Selvaraj \\
\textsuperscript{research@devan-s.com}
}
\date{May 2025}

\begin{document}
\maketitle
\begin{abstract}
    The muon optimizer has picked up much attention as of late as a possible replacement to the seemingly omnipresent Adam optimizer. Recently, care has been taken to document the scaling laws of hyper-parameters under muon such as weight decay and learning rate. However, at much larger scales the iterative orthogonalization procedure present in muon may suffer a possible issue as the singular values of random matrices shrink with scale. This paper shows this scaling behavior theoretically and empirically on random matrices but does not suggest what to do about it.
\end{abstract}

\section{Introduction}
The muon optimizer\cite{muon} has recently gained significant attention over the last few months as a promising alternative to the ubiquitous Adam\cite{adam} optimizer.  In particular, a recent paper\cite{mpretrain} has taken care to ensure the mu-P scaling laws still apply to the muon optimizer. However, the iterative orthogonalization algorithm used in muon may not scale indefinitely. Unfortunately, the distribution of the singular values does not remain constant with scale and it will possibly introduce issues when attempting to scale the muon optimizer to significantly higher parameter counts.

\textit{This paper argues for the existence of a scaling law for the coefficients and number of the NS iterations. It is not in the scope of the paper to decide the remedy.}
\newpage

\section{Iterative Orthogonalization}
Muon's preconditioner uses a clever polynomial to iteratively orthgonalize the momentum before applying it to the model. Using odd-degree exponents of the matrix, you can directly apply a polynomial to the singular values of a matrix as follows:

\begin{align*}
    G' 
    &= aG + b(GG^T)G + c(GG^T)(GG^T)G \\
    &= (aI + b(GG^T) + c(GG^T)(GG^T))G \\
    &= (aUU^T + bUS^2U^T + cUS^4U^T)USV^T \\
    &= U(a + bS^2 + cS^4)U^TUSV^T \\
    &= U(a + bS^2 + cS^4)SV^T \\
    &= U(aS + bS^3 + cS^5)V^T \\
\end{align*}

By normalizing the matrix with it's Frobenius norm, it's guaranteed that the singular values will fall between [0, 1]. Then, one can carefully choose the coefficients to approximate a function defined to be 1 within the interval. The procedure can then be applied iteratively to progressively orthogonalize the matrix.

\subsection{Muon Orthgonalization}
In muon, the coefficients of the polynomial are chosen to increase steeply at 0 to 1 to minimize the number of steps taken. However, if the expected distribution of singular values shrinks as the matrices scale, then the decisions on what constants to choose become unclear.

\begin{figure}[H]
    \centering
    \includegraphics[width=0.5\linewidth]{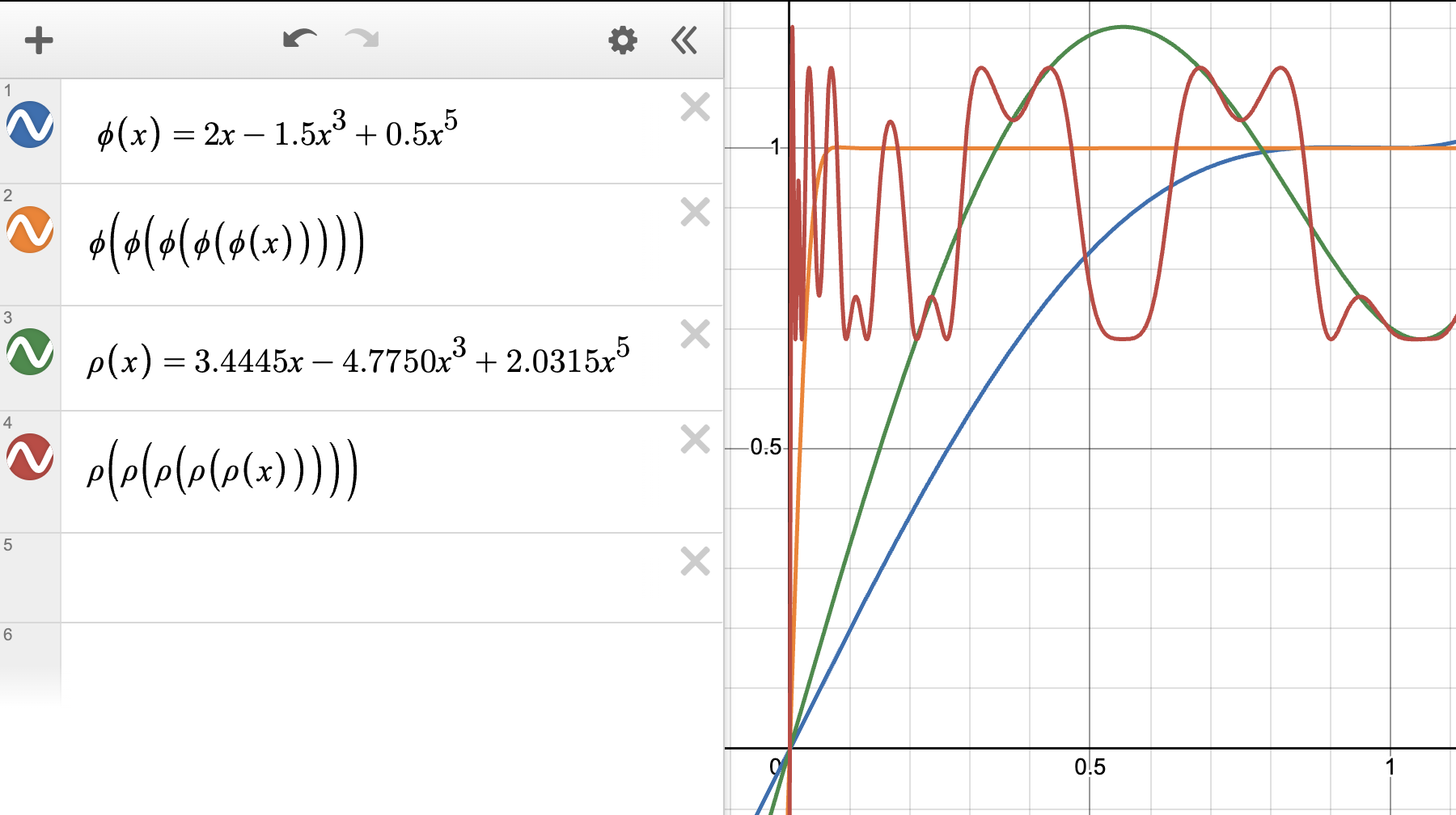}
    \caption{Muon's chosen polynomial (from Keller's Blog)\cite{muon}}
    \label{fig:enter-label}
\end{figure}

As will be shown, the expected distribution of singular values does change predictably with size. It appears their expected distribution shrinks with the square root of the smallest dimension. This causes an issue where the polynomial must be steepened or modified in order to reliably orthogonalize the matrix at larger scales.

\section{Singular Value Scaling Law}
Empirically, this scaling behavior can be observed in randomly initialized matrices. To show the distribution of singular values 8-512 matrices of size 128 through 8192 are randomly initialized $N(0, 1)$, normalized and have the singular values computed. The distributions of the singular values of the largest and smallest matrices are then plotted before and after applying the polynomial for 5 iterations.
\begin{figure}[H]
    \centering
    \includegraphics[width=.75\linewidth]{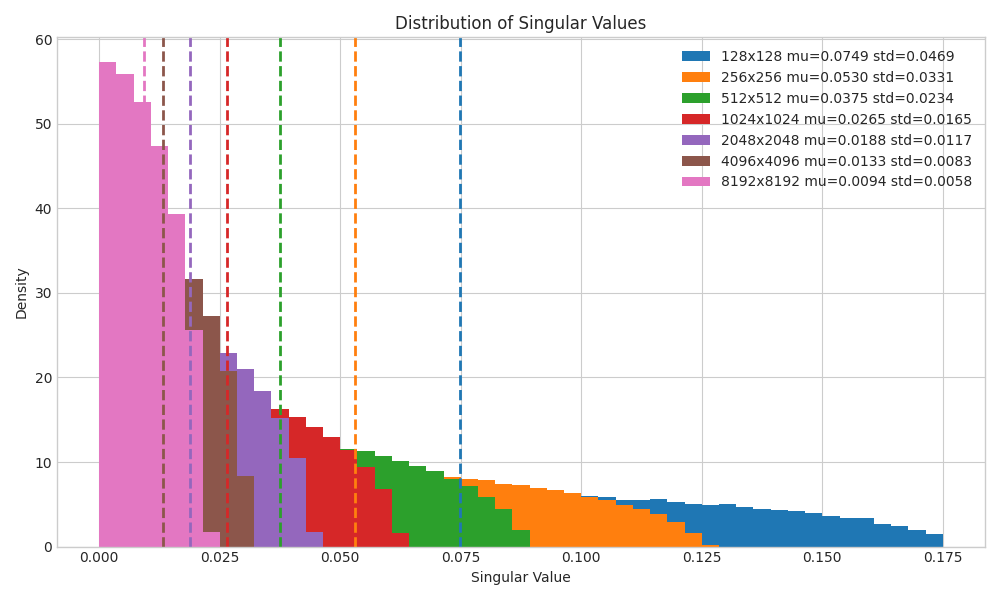}
    \vspace{.5em}
    \includegraphics[width=.75\linewidth]{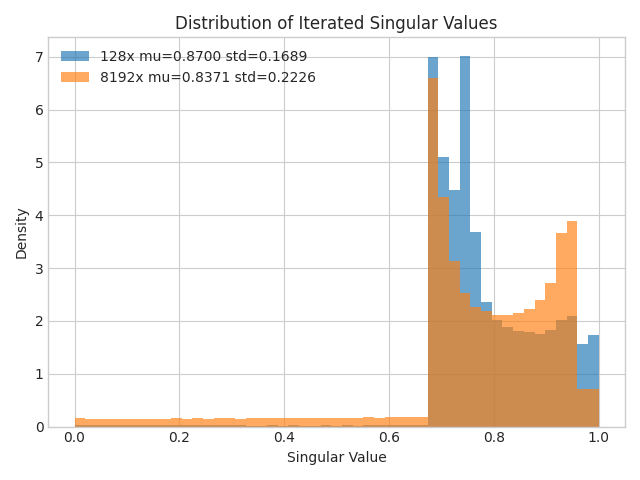}
\end{figure}
As can one can see, the higher matrix dimension has a heavier tail below .6, showing the less likely directions are not being captured by the polynomial. This issue is alleviated by increasing the number of NS iterations.


\newpage

\subsection{Proof}
The exact scaling can be calculated mathematically using the \textit{Marchenko-Pastur Theorem}\cite{marchenko}.

\begin{theorem}[Normalized Singular Value Scaling Law]\label{thm:sval-scaling-2}
    The singular values of a gradient normalized by it's Frobenius norm scale $1/\sqrt{in\_d}$.

\proof{
    \textbf{Assume} the gradient entries are i.i.d with mean $0$ finite variance $\sigma^2$. 

    \textbf{Assume} the $in\_d/out\_d$ ratio remains constant with scale

    \textbf{Apply} \textit{Frobenius Norm Scaling Law}\textsuperscript{\ref{thm:frob-scaling}} and $\nabla W$ as the gradient and $SVD_D$ is the svd of the gradient returning only the diagonals.
    \begin{align*}
        \text{SVD}_D(\dfrac{\nabla W}{||\nabla W||_F})
        &\approx \text{SVD}_D(\dfrac{N^{in\_d\times out\_d}(0, 1)}{\sqrt{in\_d \times out\_d}}) \\
        &= \dfrac{\text{SVD}_D(N^{in\_d\times out\_d}(0, 1))}{\sqrt{in\_d \times out\_d}}
    \end{align*}

    \textbf{Apply} \textit{Singular Value Scaling Law}\textsuperscript{\ref{thm:sval-scaling-1}} 
    \begin{align*}
        &\approx \dfrac{\rho \sqrt{out\_d}}{\sqrt{in\_d \times out\_d}} \\
        &= \dfrac{\rho}{\sqrt{in\_d}}
    \end{align*}
}
\end{theorem}

\section{Conclusion}
As muon continues to gain traction and larger and larger experiments are conducted using the optimizer, it is important that any possible issues are foreseen in it's scaling behavior prior to conducting experiments. The NS iteration method pioneered by muon shows promise but the current iteration requires tuning in order to capture the "\textit{unlikely directions}" that currently explain Muon's success. 

Either, the slope of the polynomial at 0 or the number of iterations should increase with scale. However, it's not clear if capturing these less likely directions is actually desirable at larger scales. LLMs have been shown to pick up outlier features on their own over around 7b parameters.\cite{int8llm} Additionally, the higher powers in the NS iterations may run into precision issues over a certain model size, but this is likely way outside realistic model sizes.

At the time of this paper, there have been no known model trains with Muon at the model dimensions necessary to cause the issues predicted in this paper. Kimi's Moonlight 16B - the largest model trained with muon so far - only had a model\_d of 2048, well within the range for Muon's default of 5 iterations to suffice. 

The issues presented in this paper will likely arise at model\_ds $>$8k, and will at worse cause less than optimal performance. 

\bibliographystyle{plain}
\bibliography{
    citations/muon,
    citations/marchenko,
    citations/mpretrain,
    citations/adam,
    citations/int8llm
}

\newpage
\section{Appendix}
\subsection{Theorems}
\begin{lemma}[Frobenius Norm Scaling Law] \label{thm:frob-scaling}
    The Frobenius norm of a matrix scales $\sqrt{out\_d \times in\_d}$ assuming constant variance.

\textbf{Assume} the elements of the gradient of a matrix are roughly normally distributed with mean $0$ and $\sigma^2$ variance.  

\begin{proof}
    \begin{align*}
        ||W||_F
        &= \sqrt{\sum_{i,j} W_{ij}^2}  \\
        &\approx \sqrt{\sum_{i,j} N(0, \sigma^2)^2}  \\
        &= \sqrt{\sigma^2 \chi^2_{in\times out}}  \\
        &\propto \sigma \sqrt{out\_d \times in\_d} \\
    \end{align*}
\end{proof}
\end{lemma}

\begin{theorem}[Marchenko-Pastur Theorem] \label{thm:marchenko-pastur}
The \textbf{Marchenko-Pastur Theorem}\cite{marchenko} states as $in\_d \rightarrow \infty$ the empirical distribution of the singular values of a matrix with i.i.d entries with $\mu=0$, variance $\sigma=\bar \sigma^2 / out\_d$, fixed aspect ratio $\gamma = out\_d/in\_d$ has singular values distributed:

    \begin{align*}
        \rho(s) &= \dfrac{\sqrt{((1+\sqrt{\gamma})^2-s^2)(s^2-|1-\sqrt{\gamma}|^2)}}{2\pi \gamma s}
    \end{align*}
\end{theorem}

\begin{lemma}[Singular Value Scaling Law]\label{thm:sval-scaling-1}
    For a matrix who's entries are i.i.d with mean 0, finite variance $\sigma^2$, and a fixed aspect ratio $\gamma = out\_d/in\_d$ the singular values scale $1/\sqrt{out\_d}$.

\begin{proof}
    \textbf{Define} initial matrix $A$
    \begin{align*}
        A &\sim N^{in\_d \times out\_d}(0, \sigma^2) \\
        &= UDV^T
    \end{align*}
    
    \textbf{Define} scaled matrix $\bar A$
    \begin{align*}
        \bar A &= A /\sqrt{out\_d} \\
        &\sim N^{in\_d \times out\_d}(0, \sigma^2/out\_d) \\
        &= U\bar DV^T \\
    \end{align*}
    
    \textbf{Apply} the \textit{Marchenko-Pastur Theorem}\textsuperscript{\ref{thm:marchenko-pastur}} the singular values are distributed by a scale-invariant distribution defined by $\rho$
    \begin{align*}
        \lim_{in\_d\rightarrow \infty} \bar D &\sim \rho
    \end{align*}
    
    \textbf{Assume} multiplying a matrix by a constant increasing it's singular values linearly, $D$ scales with $out\_d$
    \begin{align*}
        \bar A &= A /\sqrt{out\_d} \\
        U\bar DV^T &= UDV^T /\sqrt{out\_d} \\
        \bar D &= D /\sqrt{out\_d}\\
        D / \sqrt{out\_d} &\sim \rho \\
        D &\sim \rho \sqrt{out\_d}\\
        &\propto \sqrt{out\_d}
    \end{align*}
\end{proof}

\end{lemma}
\end{document}